\documentclass[letterpaper, 10 pt, conference]{ieeeconf}  

\usepackage[colorlinks,urlcolor=blue,linkcolor=blue,citecolor=blue]{hyperref}
\usepackage{color,array}
\usepackage{graphicx}
\usepackage{mathptmx}
\usepackage{algorithm}
\makeatother
\usepackage{mathtools}
\usepackage[utf8]{inputenc}
\usepackage[table]{xcolor}
\usepackage{tabularx,booktabs}
\newcolumntype{C}{>{\centering\arraybackslash}X} 
\usepackage{cite}
\usepackage{amsmath,amssymb,amsfonts}
\usepackage{graphicx}
\usepackage{textcomp}
\usepackage{xcolor}
\usepackage[mathscr]{euscript}
\usepackage{bm}
\usepackage{newtxtext,newtxmath}

\usepackage{multirow,booktabs}
\usepackage{makecell}
\usepackage{paralist}

\definecolor{MYBLUE}{RGB}{66,164,188}
\definecolor{MYORANGE}{RGB}{240,140,73}
\definecolor{MYGREEN}{RGB}{75,144,102}

\def\BibTeX{{\rm B\kern-.05em{\sc i\kern-.025em b}\kern-.08em
    T\kern-.1667em\lower.7ex\hbox{E}\kern-.125emX}}

\IEEEoverridecommandlockouts                              

\title{\LARGE \bf
Policies over Poses: Reinforcement Learning based Distributed Pose-Graph Optimization for Multi-Robot SLAM}
\author{Sai Krishna Ghanta \and Ramviyas Parasuraman 
\thanks{School of Computing, University of Georgia, Athens, GA 30602, USA.}
\thanks{Author emails: {\fontfamily{qcr}\selectfont \{sai.krishna;ramviyas\}@uga.edu}}
\thanks{This research was supported by the Army Research Laboratory and was accomplished under DCIST Cooperative Agreement W911NF-17-2-0181. }
}

\begin{document}
\maketitle

\thispagestyle{empty}
\pagestyle{empty}
\newcommand{\revision}{\textcolor{red}}

\begin{abstract}
We consider the distributed pose-graph optimization (PGO) problem, which is fundamental in accurate trajectory estimation in multi-robot simultaneous localization and mapping (SLAM). Conventional iterative approaches linearize a highly non-convex optimization objective, requiring repeated solving of normal equations, which often converge to local minima and thus produce suboptimal estimates. We propose a scalable, outlier-robust distributed planar PGO framework using Multi-Agent Reinforcement Learning (MARL). We cast distributed PGO as a partially observable Markov game defined on local pose-graphs, where each action refines a single edge’s pose estimate. A graph partitioner decomposes the global pose graph, and each robot runs a recurrent edge-conditioned Graph Neural Network (GNN) encoder with adaptive edge-gating to denoise noisy edges. Robots sequentially refine poses through a hybrid policy that utilizes prior action memory and graph embeddings. After local graph correction, a consensus scheme reconciles inter-robot disagreements to produce a globally consistent estimate. Our extensive evaluations on a comprehensive suite of synthetic and real-world datasets demonstrate that our learned MARL-based actors reduce the global objective by an average of 37.5\% more than the state-of-the-art distributed PGO framework, while enhancing inference efficiency by at least 6X. We also demonstrate that actor replication allows a single learned policy to scale effortlessly to substantially larger robot teams without any retraining. Code is publicly available at \url{https://github.com/herolab-uga/policies-over-poses}
\end{abstract}

\section{Introduction}
Collaborative Simultaneous Localization and Mapping (C-SLAM) \cite{fernandez2024multi,tian2022kimera,ghanta2024space} enables a multi-robot system to estimate their trajectories cooperatively and fuse local observations into a globally consistent map, thereby boosting the autonomy and cognitive capability of the multi-robot system. Distributed C-SLAM backend optimization is driven by distributed PGO, a fundamental mathematical abstraction in which each robot solves a maximum-likelihood, least-squares problem that minimizes the mismatch between noisy measurements such as odometry, loop closures, and relative estimates \cite{tian2021distributed,li2024distributed,ghanta2025mgprl,latif2022dgorl}.

However, computing maximum likelihood estimates for robot poses in distributed PGO presents substantial challenges due to the inherent complexity of solving high-dimensional, non-convex optimization problems characterized by multiple local minima \cite{tian2020asynchronous}. Existing C-SLAM approaches frequently involve highly non-convex cost functions due to factors such as system nonlinearity, substantial measurement noise, large distances between connected poses, and incorrect data associations originating from the front-end processes \cite{carlone2014fast}. These complexities manifest as broad, flat regions in the optimization landscape, hindering traditional gradient-based optimization methods sensitive to initial estimates \cite{olson2006fast}. Consequently, classical approaches often fail to achieve global optimal solutions in time and instead converge to suboptimal or unsatisfactory outcomes.

\begin{figure}[t]
    \centering
    \includegraphics[width = \linewidth]{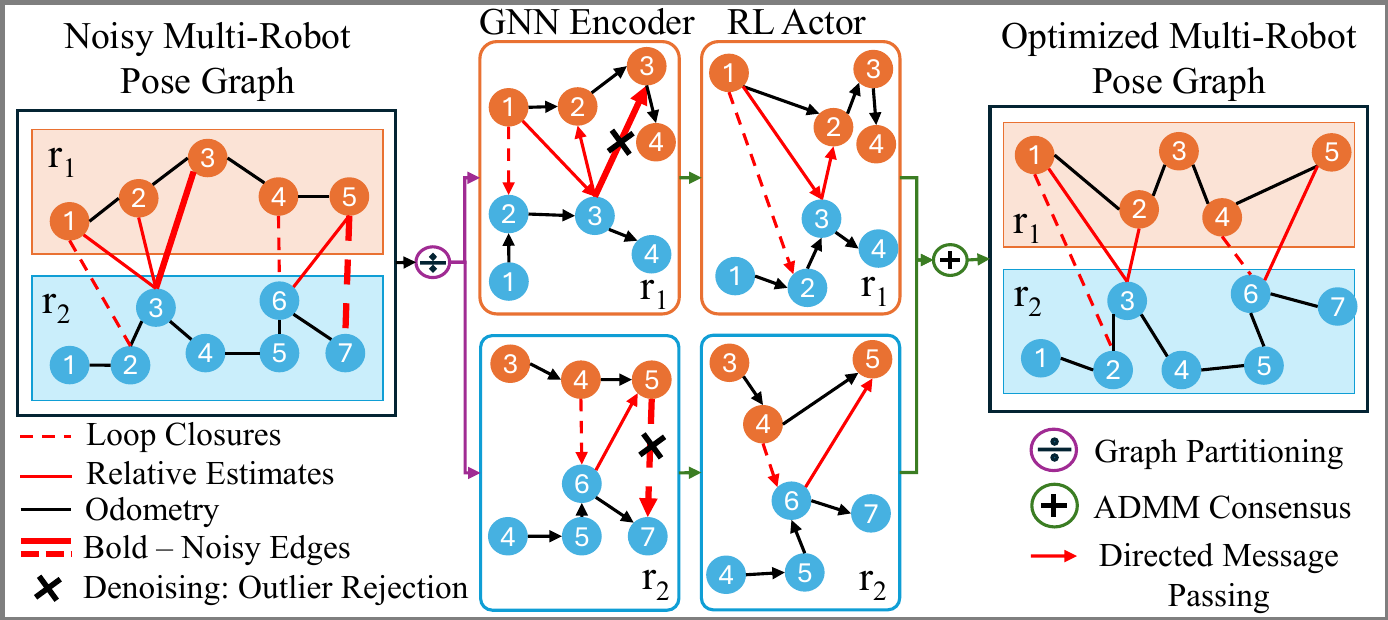}
    \vspace{-6mm}
    \caption{\footnotesize Illustration of the proposed distributed framework. Each robot processes its partitioned subgraph with a GNN encoder equipped with a denoising mechanism, iteratively updates edge poses, and a final consensus step reconciles overlaps into a globally consistent pose graph.}
    \label{fig:intro_1}
    \vspace*{-7mm}
\end{figure}

Although limited in number, learning-based graph optimization methods, particularly those using Reinforcement Learning (RL) \cite{kourtzanidis2023rl} and Graph Neural Networks (GNN) \cite{nejatishahidin2023graph,azzam2021pose}, show promise in generating accurate initial estimates that serve as informed priors to improve convergence and reduce sensitivity to initialization in classical solvers. However, these works focus solely on estimating rotation components while ignoring translations, which results in partial solutions that fail to capture the full geometric structure of the pose graph, ultimately limiting applicability for complete pose graph optimization. Furthermore, these methods \cite{kourtzanidis2023rl} lack dedicated denoising mechanisms, leaving them vulnerable to outlier measurements that degrade the optimization quality. Critically, these approaches are designed for single-robot trajectories, making them incompatible with distributed systems where robots operate under noisy and partial observability. 

Therefore, we introduce a distributed PGO framework based on Graph-Aware Soft Actor Critic (GA-SAC) network that decisively tackles three bottlenecks of prior works—accurate translational and rotational refinement, learnable denoising, and scalability to large robot teams. Each robot runs an edge-conditioned GNN encoder equipped with an adaptive edge-gating denoiser that suppresses outlier measurements. The MARL actor then applies edge-wise pose corrections, as illustrated in Fig. \ref{fig:intro_1}. The core contributions of this paper are that
    (1) we cast multi-robot PGO as a cooperative, partially observable Markov game in which each robot optimizes its local subgraph and iteratively reaches consensus on the global pose graph;
    (2) we propose an edge-conditioned GNN encoder with adaptive edge-gating that simultaneously denoises corrupted constraints and then drives a sequential edge selection and pose correction policy.
Extensive evaluations on large-scale datasets show that the proposed framework outperforms existing distributed PGOs and generalizes across graph topologies.

\section{Literature Review}
\subsection{Distributed PGO in SLAM}
The existing distributed PGO algorithms almost universally adopt the classical linearize-solve-iterate solvers such as g\textsuperscript{2}o \cite{kummerle2011g} and GTSAM \cite{dellaert2012factor}. They repeatedly linearise the non-convex maximum-likelihood objective, solve sparse normal equations with either a Gauss–Newton (GN) step or its damped variant, Levenberg–Marquardt (LM), and exchange intermediate estimates across the multi-robot team. However, their convergence to the global optimum critically depends on the quality of the initial guess \cite{carlone2014fast}, as poor guesses frequently cause the solver to stall in suboptimal local minima. Tian et al \cite{tian2021distributed}\cite{tian2020asynchronous} introduced Certifiably Correct PGO (CC-PGO), which offers global-optimality guarantees only if it is seeded with sufficiently accurate rotation guesses. To mitigate the sensitivity to initialization, Fan et al. \cite{fan2023majorization} introduced a majorization–minimization (MM-PGO) scheme boot-strapped by chordal synchronization, yet their method still presumes a near-accurate rotational seed and degrades sharply under large drift due to heuristic robust kernels for outlier rejection. Moreover, these state-of-the-art distributed solvers \cite{li2024distributed} \cite{tian2020asynchronous}\cite{fan2023majorization} often require hundreds of iterations to converge on noisy graphs, motivating learning-based methodologies that can perform online denoising, amortize computation, and provide policy-driven warm starts for large-scale, real-time deployment.

Recent advances has embedded learning-based optimizers directly into SLAM--e.g. DeepV2D’s neural bundle-adjustment core \cite{teed2018deepv2d} and BA-Net’s learned photometric optimizer \cite{tang2018ba}. Extending these ideas, end-to-end factor-graph smoothers trained via surrogate losses~\cite{yi2021differentiable} now achieve state-of-the-art performance on visual tracking and odometry.

\subsection{Graph Neural Networks (GNNs)}
GNNs offer equivariant message-passing that naturally suits geometric inference, and have yielded state-of-the-art results across visual domains \cite{jiang2019semi, shi2020message}. In graph settings, NeuRoRA \cite{purkait2020neurora} pioneered an end-to-end strategy for multiple-rotation averaging (MRA) by stacking a message passing neural network (MPNN) with a subsequent regressor, while Thorpe et al.\ \cite{thorpe2020rotation} showed that a single edge-attention MPNN can match NeuRoRA’s accuracy at half the training cost and one-quarter the inference time. Moreover, PoGO-Net \cite{li2021pogo} presents a joint-loss formulation for MRA and learnable denoising that surpasses previous state-of-the-art results while still running in real-time. However, the straightforward adoption of these GNNs for PGO fails because uniform message weights treat all edges as equally reliable, ignoring the heterogeneity of odometry, loop closures, and inter-robot estimates.

RL-PGO \cite{kourtzanidis2023rl} marks the first effort to pose-graph optimization as a single-agent RL task using a GNN architecture. The formulation of this method is fundamentally constrained: the learned policy corrects only rotational variables, disregarding translations and therefore omitting the coupled geometry. The method provides no explicit denoising, leaving it susceptible to outliers, and it is trained exclusively on graphs of 20 poses, limiting generalizability to larger graphs encountered in multi-robot deployments. Moreover, RL-PGO assigns uniform importance to all weights, treating odometry and loop-closure edges identically despite their distinct noise characteristics and topological influence. These limitations motivate the present work’s to consider translations in action space, adaptive denoising, and scalability to graphs orders of magnitude larger.

\subsection{Multi-Agent Reinforcement Learning (MARL)}
Predominant MARL frameworks adopt the centralised training decentralized execution (CTDE) paradigm, in which a critic learns from the state–action space while each agent executes a lightweight, local policy \cite{lan2024multiagent}. The multi-agent extensions of Soft Actor–Critic (SAC), such as MA-SAC \cite{pu2021decomposed} and FAC-SAC \cite{yue2023factored}, are especially attractive because entropy-regularized updates yield superior sample efficiency and robustness. However, direct adaptation of these existing MARL SAC frameworks is infeasible because the PGO task demands hybrid actions (edge selection and pose correction) and graph-structured observations. 

At inference time, all three distributed frameworks—CC-PGO \cite{tian2021distributed}, ML-PGO \cite{li2024distributed}, and MM-PGO \cite{fan2023majorization}—remain highly sensitive to the initial pose estimate and generally require on the order of $10^2-10^3$ iterations for convergence. CC-PGO and ML-PGO lack robust outlier‐rejection mechanisms, whereas MM-PGO applies a heuristic kernel filter whose fixed break-points requires tedious tuning for each dataset. The proposed MARL framework addresses these limitations by learning lightweight policies that produce near-optimal pose estimates and maintain accuracy under extreme noise. Moreover, these near-optimal estimates can be fed into classical solvers as initializations, requiring only on the order of tens of iterations to converge to the global optimum. Existing GNN-based approaches further falter by assigning uniform trust to heterogeneous constraints. We address this with a modality-aware, edge-conditioned GNN equipped with adaptive edge-gate denoising, whose learned weights systematically down-weight noisy edges. RL-PGO \cite{kourtzanidis2023rl} scales poorly, correcting only rotations on single-robot small graphs. By training distributed actors on large-scale datasets, we generalize translational–rotational edge corrections to full $\mathbb{SE}(2)$ pose graphs.

\section{Methodology}
\begin{figure}[t]
    \centering
    \includegraphics[width = \linewidth]{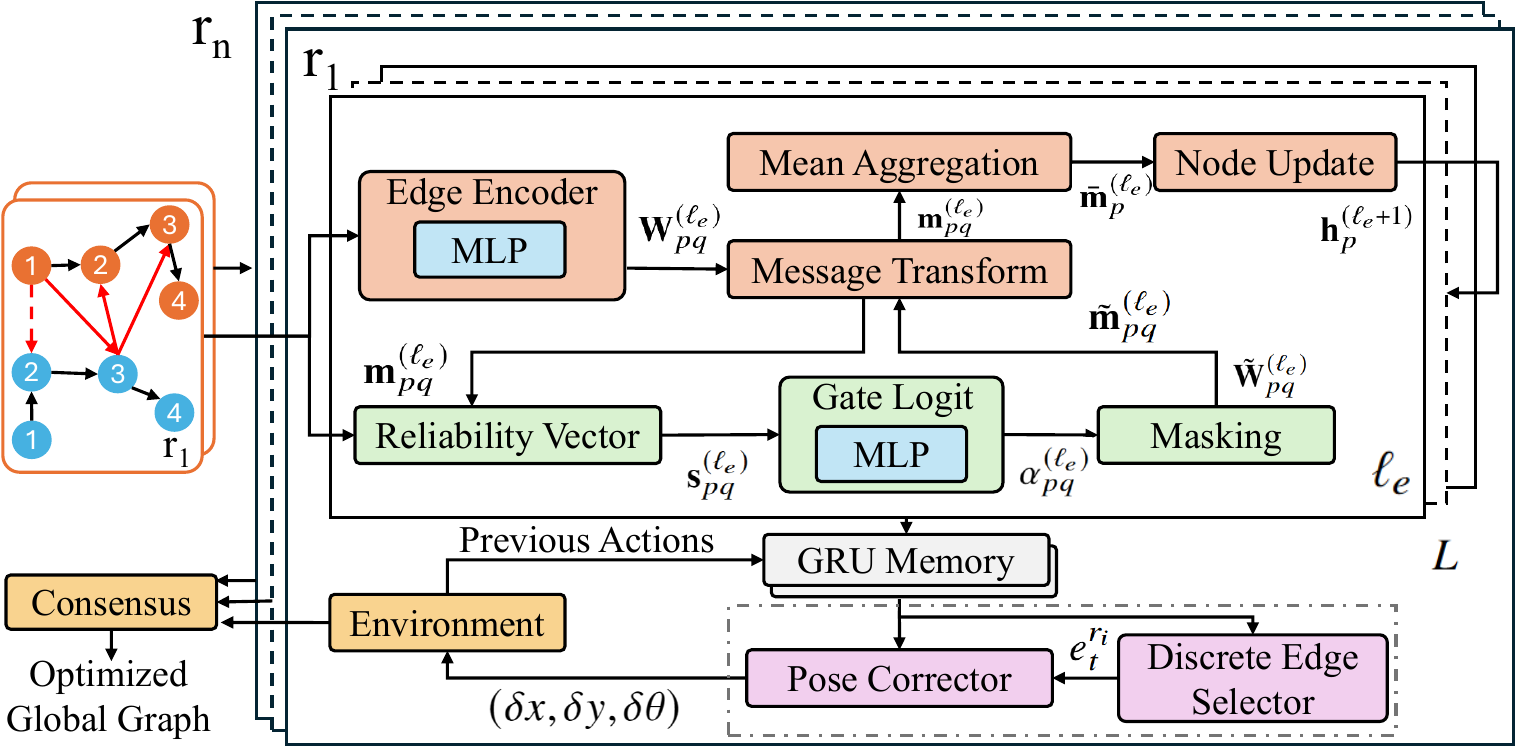}
    \vspace{-6mm}
    \caption{\footnotesize{Overview of the proposed methodology. The \textcolor{MYORANGE}{-orange} represents \textbf{edge-conditioned GNN encoder}, \textcolor{MYGREEN}{-green} denotes \textbf{adaptive edge-gate denoising}, and \textcolor{violet}{-violet} indicates the action heads in \textbf{MARL actor}.}}
    \label{fig:pm}
    \vspace*{-4mm}
\end{figure}

The proposed approach is a \emph{distributed framework for scalable, outlier-robust multi-robot} PGO as illustrated in Fig.~\ref{fig:pm}. Given a noisy global pose graph, a multi-level graph partitioner \cite{li2024distributed} decomposes the global pose graph into fine-scale local robot graphs assigned to individual robots, while preserving the inter-robot graph dependencies. Each robot processes its local partition with an edge-conditioned GNN encoder equipped with an adaptive edge–gate denoising mechanism that suppresses corrupted constraints and yields a compact latent state. This latent state, concatenated with the previously selected edge and correction, updates a replayable GRU memory that maintains temporal context for sequential decision making. This refreshed memory is fed to two heads at every decision step—(i) discrete edge selector and (ii) a continuous pose corrector. During training, the distributed actors share experience with a central critic under a CTDE paradigm: the critic ingests the concatenated latent graphs and actions of all robots and is optimized with a GA-SAC objective, whereas at inference, only the actors execute autonomously. Once all local pose corrections have been made, the global pose graph is harmonised through an information-weighted ADMM \cite{fung2019uncertainty} that reconciles inter-robot constraints.

\subsection{Problem Formulation}
Consider a team of $n$ robots $\mathcal{R}=\{r_{1},\dots,r_{n}\}$ operating in a planar workspace, under the ideal communication with no latency. The objective of PGO is to determine the most consistent set of robot poses given a collection of noisy relative measurements. We represent the problem by a directed pose graph $\mathcal{G}=(\mathcal{V},\mathcal{E})$, where each vertex $v_{\tau}^{i}\!\in\!\mathcal{V}$ encodes the unknown planar pose
$\mathbf{x}_{i}^{\tau}\!=\!(\mathbf{R}_{i}^{\tau},\mathbf{t}_{i}^{\tau})\!\in\!\mathbb{SE}(2)$
of robot $r_i$ at discrete timestep ~$\tau\!\in\!\mathbb{N}$.  
Every directed edge $(v_{\tau}^{p},v_{s}^{q})\!\in\!\mathcal{E}$ stores the relative transformation
$\tilde{\mathbf{x}}_{p_{\tau}}^{\,q_{s}}\!=\!(\tilde{\mathbf{R}}_{p_{\tau}}^{\,q_{s}},\tilde{\mathbf{t}}_{p_{\tau}}^{\,q_{s}})\!\in\!\mathbb{SE}(2)$
observed from $v_{\tau}^{p}$ to $v_{s}^{q}$, its information matrix
$\boldsymbol{\Lambda}_{p_{\tau}}^{\,q_{s}}\!\in\! \mathbb{S}{++}(3)$, and an origin label ($\textbf{O}_p^{q}$) distinguishing odometry, relative, and loop-closure constraints. 
A graph partitioner decomposes $\mathcal{G}$ into $n$ disjoint subgraphs
$\{\mathcal{G}^{\,r_i}\}_{i=1}^{n}$; robot $r_i$ receives
$\mathcal{G}^{\,r_i}=(\mathcal{V}^{r_i},\mathcal{E}^{r_i})$. We formulate the distributed global PGO optimization $F(x)$ objective as
\vspace{-1mm}
\begin{equation}
\begin{aligned}
\min
\; \sum_{(v_{\tau}^{p},v_{s}^{q})\!\in\!\mathcal{E}}
\Bigl(
    w_R^{2}\,
    \bigl\|
        \operatorname{Log}\!\bigl(
            \tilde{\mathbf{R}}_{pq}^{\top}\mathbf{R}_p^{\top}\mathbf{R}_{q}
        \bigr)
    \bigr\|^{2}
    + \\
    w_T^{2}\,
    \bigl\|
        \mathbf{R}_p^{\top}(\mathbf{t}_q-\mathbf{t}_p)
        -\tilde{\mathbf{t}}_{pq}
    \bigr\|_{2}^{2}
\Bigr) ,
\vspace{-1mm}
\label{eq:local_pgo}
\end{aligned}
\end{equation}
where $\mathbf{R}_{p},\mathbf{R}_{q}\!\in\!\mathbb{SO}(2)$ are rotation matrices,  
$\mathbf{t}_{p},\mathbf{t}_{q}\!\in\!\mathbb{R}^{2}$ are planar translations,  
$\tilde{\mathbf{R}}_{pq},\tilde{\mathbf{t}}_{pq}$ are the measured relative rotation and translation,  
$\texttt{Log}(\cdot)$ denotes the matrix logarithm mapping, 
$\|\cdot\|_{2}$ the Euclidean norm, and  
$w_{R},w_{T}\!>\!0$ are weighting factors balancing rotational versus translational residuals.  

\subsection{Environment}
\textbf{Sequential Decision-Making:} We formulate multi-agent PGO as a finite-horizon Markov game whose hidden state is the global pose-graph estimate, each agent’s observation is its local subgraph, each step’s joint action selects incident edges and performs pose refinement.  This edge-wise decomposition collapses the control space to a discrete edge index. Each robot selects a single directed edge in its local graph and proposes a small pose correction until all corrections are made. At each discrete timestep $t\in \{0,1....T-1\}$, all robots simultaneously perform observations and actions, then receive individual rewards once the environment processes the joint action and updates the underlying graph. The episode ends once every original edge has been refined, giving a fixed horizon $T=|\mathcal{E}|$. The episodes, therefore, have a fixed and known horizon determined by the number of edges in the pose graph.

\vspace{2mm}
\textbf{Observation Space:} Robot \(r_i\) observes the directed local graph
\(
\mathcal{G}^{\,r_i}_t=(\mathcal{V}^{r_i},\mathcal{E}^{r_i}_t)
\).
Because each step refines exactly one unprocessed edge, the
edge set evolves deterministically:
$
\mathcal{E}_{t+1}
\;=\;
\mathcal{E}_{t}\setminus\bigl\{e_t^{\,r_1},\dots,e_t^{\,r_n}\bigr\}.
$
Hence the state space shrinks monotonically until
\(\mathcal{E}_{T}=\varnothing\).

\vspace{2mm}
\textbf{Action Space:} The agent’s action space is a hybrid discrete–continuous tuple $a_t^{r_i} = \bigl(e_t^{r_i},\;\Delta_t^{r_i}\bigr)\in
\mathcal{A}^{r_i} $, where $e_t^{r_i}$ selects one edge within the agent’s local partition to refine; $\Delta_t^{r_i}=(\delta x,\delta y,\delta\theta)$ is a bounded pose correction for the edge.\footnote{We enforce a safety constraint on each pose correction by limiting its magnitude to $\Delta_{max}^{r_i}$ prior to application.}
        
\vspace{2mm}
\textbf{Reward:} Let $\mathcal{L}(\mathcal{G})$ denote the \emph{localization $\chi^2$ error} of a pose-graph $\mathcal{G}$ calculated from ground truth and corrected measurements. Instead of rewarding every raw reduction of the localized $\chi^{2}$ loss, we densely reward the agent \emph{only for the fraction of error it removes relative to the error that was still left}.  After each action we compute the normalized gain as:
\begin{equation}
    R_t \;=\; \frac{\mathcal{L}(\mathcal{G}_{t-1}) - \mathcal{L}(\mathcal{G}_t)}{\mathcal{L}(\mathcal{G}_{t-1}) + \varepsilon}
\end{equation}
, with a tiny $\varepsilon$ to avoid division by zero.  
The per-step reward is then smoothed via \texttt{Tanh} and clipped.
At the final step we give a bonus proportional to the overall
error ratio $\texttt{Log}(\mathcal{L}(\mathcal{G}_0)/(\mathcal{L}(\mathcal{G}_T)+\varepsilon))$, so the agent is judged both on steady progress and the quality of the final pose graph.  

\vspace{-1mm}
\subsection{Graph Neural Network}
Graph Neural Networks (GNNs) are proven to capture high-order geometric correlations while respecting the sparsity pattern of $\mathcal{G}$ \cite{li2021pogo,kourtzanidis2023rl}. The existing RL-PGO approach \cite{kourtzanidis2023rl} treats all edges uniformly in their GNN encoder, which means loop closures and noisy odometry edges influence the estimate equally. To tackle this problem, inspired by \cite{simonovsky2017dynamic}, we propose an edge-conditioned convolution with normalized mean aggregation. Furthermore, RL-PGO \cite{kourtzanidis2023rl} offers no explicit denoising module to suppress corrupted measurements—a shortcoming the authors themselves identify as future work. We close this gap with an adaptive edge-gate denoising mechanism that automatically detects and prunes spurious edge constraints.

\textbf{Message Aggregation:}
For layers $\ell = 0,\dots,L-1$ with dimension $d_{l}$, let
$\mathbf{h}^{(\ell)}_p \!\in\! \mathbb{R}^{d_{\ell}}$ be the node embedding of
vertex $v_p$. Each directed edge $(v_{\tau}^{p},v_{s}^{q})\!\in\!\mathcal{E}$ is annotated with an attribute vector $\mathbf{a}_{pq}\!\in\!\mathbb{R}^{d_e}$ that concatenates the measurement-origin label
$\textbf{O}_p^{q}$, the log-information matrix
$\log {\boldsymbol{\Lambda}_{p}^{q}}^{-1}$, the timestep separation ($\tau - s\!\ge\!0$), the translation magnitude
$\lVert \tilde{\mathbf{t}}_{pq} \rVert_2$, and a wrap-free angular encoding
$[\sin\tilde{\theta}_{pq},\,\cos\tilde{\theta}_{pq}]$. The layer-wise update at layer $\ell = \ell_e$ is as follows:
\begin{align}
\mathbf{W}^{(\ell_e)}_{pq}
  &= \phi^{(\ell_e)}_{\text{edge}}\!\bigl(\mathbf{a}_{pq}\bigr)
     &&\in \mathbb{R}^{d_{\ell_e+1}\times d_{\ell_e}},                    \label{eq:ecc_weight}\\[2pt]
\mathbf{m}^{(\ell_e)}_{pq}
  &= \mathbf{W}^{(\ell_e)}_{pq}\,\mathbf{h}^{(\ell_e)}_q
     &&\in \mathbb{R}^{d_{\ell_e+1}},                                     \label{eq:ecc_message}\\[2pt]
\bar{\mathbf{m}}^{(\ell_e)}_p
  &= \frac{1}{\lvert \mathcal{N}(i) \rvert}\!
     \sum_{q\in\mathcal{N}(p)} \mathbf{m}^{(\ell_e)}_{pq}
     &&\in \mathbb{R}^{d_{\ell_e+1}},                                     \label{eq:mean_aggr}\\[2pt]
\mathbf{h}^{(\ell_e+1)}_p
  &= \sigma\!\Bigl(
       \mathbf{U}^{(\ell_e)} \mathbf{h}^{(\ell_e)}_p
       \;\bigl\|\;
       \bar{\mathbf{m}}^{(\ell_e)}_p
     \Bigr)
     &&\in \mathbb{R}^{d_{\ell_e+1}},                                     \label{eq:node_update}
\end{align}
where the edge attribute vector $\mathbf{a}_{pq}$ is processed by a learnable MLP network $\phi^{(\ell_e)}_{\text{edge}}$, yielding the edge-conditioned weight matrix $\mathbf{W}^{(\ell_e)}_{pq}$ in~\eqref{eq:ecc_weight}.  
This matrix linearly projects the node embedding $\mathbf{h}^{(\ell_e)}_{q}$ to produce the directed message $\mathbf{m}^{(\ell_e)}_{pq}$ defined in~\eqref{eq:ecc_message}.  
All incoming messages at node $p$ are aggregated with a degree-normalized mean as in~\eqref{eq:mean_aggr}, thereby avoiding bias toward high-degree vertices. The aggregated message is then fused with the node’s self-embedding, transformed by a learnable matrix $\mathbf{U}^{(\ell_e)}$, passed through the non-linearity function $\sigma$ (logistic sigmoid), yielding next-layer embedding $\mathbf{h}^{(\ell_e+1)}_{p}$ in \eqref{eq:node_update}. This differentiable pipeline
$(\mathbf{a}_{pq},\mathbf{h}^{(\ell_e)}_{q})\!\mapsto\!\mathbf{m}^{(\ell_e)}_{pq}$ pushes the filters of uninformative edges toward a near-zero spectral norm. Although the final node embeddings $\mathbf{h}^{(L)}_p$ capture the current topological consistency of the pose graph, our policy acts sequentially on individual edges.
We furnish the policy with a memory comprising $K$ stacked gated-recurrent units (GRUs), refreshed after every processed edge to retain the context accumulated over the ordered stream of edge updates.


\vspace{2mm}
\textbf{Adaptive Edge–Gate Denoising:}
After each message-passing layer the graph is denoised by a lightweight \emph{adaptive edge gate} (AEG) that prunes unreliable constraints while remaining fully differentiable.  Let $\mathbf{r}_{pq}^{(\ell_e)}\!\in\!\mathbb{R}^{3}$ be the pose residual in minimal log–map coordinates between the current estimate and the raw measurement $\tilde{\mathbf{x}}_{pq}$,\footnote{The actor emits a single pose update; the residual $\mathbf{r}_{pq}^{(\ell_e)}$ is then evaluated on this provisional graph, so informative consistency cues are still available in single-shot inference.} and let $\mathbf{c}_{pq}=\texttt{vec}\!\bigl(\log\boldsymbol{\Lambda}_{p}^q\bigr)\!\in\!\mathbb{R}^{3}$ encode the log-covariance vector.  Concatenating these cues with the directed message yields the reliability vector:
\begin{equation}
\mathbf{s}_{pq}^{(\ell_e)}=[\,\mathbf{m}_{pq}^{(\ell_e)}\|\mathbf{r}_{pq}^{(\ell_e)}\|\mathbf{c}_{pq}\,]\in\mathbb{R}^{d_{\ell_e+1}+6},
\label{eq:reliability_vector}
\end{equation}

which is further mapped by MLP $\mathcal{P}_{\text{gate}}^{(\ell_e)}$ to extract a scalar gate logit/ edge confidence:
\begin{equation}
    \vspace{-1mm}
    \alpha_{pq}^{(\ell_e)}=\mathcal{P}_{\text{gate}}^{(\ell_e)}\!\bigl(\mathbf{s}_{pq}^{(\ell_e)}\bigr)
   \;+\;
   \beta\,,
\label{eq:logit}
\end{equation}

where a larger value indicates higher confidence that edge $(v^p, v^q)$ is reliable.  A small bias $\beta\!>\!0$ is added whenever the measurement label $\textbf{O}_p^{q}$ denotes an inter-robot loop closure, promoting the retention of long-range loop closures.  To keep the edge gate differentiable and learnable, the logit
$\alpha_{pq}^{(\ell_e)}$ is passed through the Concrete
(Gumbel–Softmax) relaxation \cite{maddison2016concrete}, producing a soft edge mask. A relaxed Bernoulli sample is then drawn as:
\begin{equation}
    u_{pq}^{(\ell_e)}
      =\sigma\!\Bigl(\tfrac{\alpha_{pq}^{(\ell_e)}+\log\epsilon-\log(1-\epsilon)}{\tau}\Bigr),
      \qquad
      \epsilon\sim\mathcal{U}(0,1),
\label{eq:bernoulli}
\end{equation}
where $u_{pq}^{(\ell_e)}\!\in\!(0,1)$ carries gradient information.
Stretching and clipping to $(a,b)$ yields the final gate:
\begin{equation}
z_{pq}^{(\ell_e)}
      =\texttt{min} \!\bigl(1,\texttt{max}\bigl((b-a)u_{pq}^{(\ell_e)}+a,0\bigr)\bigr),
    \label{eq:minmax}
\end{equation}

with $(a,b)=(-0.1,1.1)$. According to the hard-concrete scheme of \cite{louizos2017learning}, choosing bounds slightly outside $[0,1]$ prevents early saturation at the clipping thresholds, so gradients remain informative. The mask multiplies both the edge weight and the routed message:
\begin{equation}
    \tilde{\mathbf{W}}_{pq}^{(\ell_e)}=z_{pq}^{(\ell_e)}\mathbf{W}_{pq}^{(\ell_e)},
\qquad
\tilde{\mathbf{m}}_{pq}^{(\ell_e)}=\tilde{\mathbf{W}}_{pq}^{(\ell_e)}\mathbf{h}_{q}^{(\ell_e)},
\label{eq:mask}
\end{equation}

so edges with $z_{pq}^{(\ell_e)}\!\approx\!0$ contribute negligibly to aggregation.  
An $\mathcal{L}_{1}$ penalty on  $z_{pq}^{(\ell_e)}$, is performed to encourage sparsity, driving noisy or redundant constraints toward zero. During inference time, we can simply threshold $z_{pq}^{(\ell_e)}$ to obtain a pruned pose-graph that preserves high-quality odometry and loop closures while suppressing outliers.

\subsection{Graph-Aware Multi-Agent Soft Actor-Critic (GA--SAC)}
\label{sec:ga_sac}
Most off-the-shelf multi–agent SAC variants \cite{hu2023graph, wang2024multi, li2021structured} struggle with \emph{distributed pose–graph optimization} (PGO) because they  
(i) support only purely continuous or discrete actions, whereas PGO requires the
hybrid choice of which edge to refine (discrete) and a small
pose correction (continuous); and  (ii) build value functions over the full Cartesian product of all agents’
actions, incurring a prohibitive $\mathcal{O}(n^{2}|\mathcal{E}|)$ complexity. Our GA--SAC extends the original multi-agent SAC formulation \cite{iqbal2019actor} with two modifications tailored for PGO:
\begin{enumerate}
    \item \textbf{Distributed graph actors:}  
          Each robot $r_i\!\in\!\mathcal{R}$ encodes its local partition
          $G^{\,r_i}_t$ with a recurrent edge-conditioned GNN, producing a latent
          vector $\mathbf{h}^{\,r_i}_t\!\in\!\mathbb{R}^d$.
          The actor
          $\pi_{\psi^{r_i}}$ comprises  
          (i) a head that samples an edge index and  
          (ii) a head that outputs
          $\Delta_t^{\,r_i}=(\delta x,\delta y,\delta\theta)$.
    \item \textbf{Entropy-regularised CTDE critic:}  
          During training, a central critic embeds every robot’s graph with a
          shared non-recurrent GNN, concatenates all latent vectors,
          chosen edges, and pose corrections, and evaluates the soft
          state–action value $Q_{\varphi}(\mathbf{s}_t,\mathbf{a}_t)$.
\end{enumerate}

\begin{table*}[t]
  \centering
  \caption{Performance comparison against state-of-the-art benchmarks on standard real-world datasets for varying team sizes}
  \label{tab:ale_time_benchmarks}
  \vspace{-2mm}
  \scriptsize
  \setlength{\tabcolsep}{2pt}
  \renewcommand{\arraystretch}{1.05}
  \resizebox{\textwidth}{!}{%
  \begin{tabular}{|c|c|ccccc|ccccc|ccccc|}
    \hline
    \multirow{2}{*}{\makecell{\textbf{Dataset} \\ \footnotesize(\# poses, \# edges)}} &
    \multirow{2}{*}{\textbf{Metric}} &
    \multicolumn{5}{c|}{\textbf{Number of Robots $n = 3$}} &
    \multicolumn{5}{c|}{\textbf{$n = 7$}} &
    \multicolumn{5}{c|}{\textbf{$n = 35$}} \\ \cline{3-17}
    & & \textbf{CC-PGO} & \textbf{ML-PGO} & \textbf{MM-PGO} & \textbf{Prop-V1} & \textbf{Prop-V2}%
      & \textbf{CC-PGO} & \textbf{ML-PGO} & \textbf{MM-PGO} & \textbf{Prop-V1} & \textbf{Prop-V2}%
      & \textbf{CC-PGO} & \textbf{ML-PGO} & \textbf{MM-PGO} & \textbf{Prop-V1} & \textbf{Prop-V2} \\ \hline

    \multirow{2}{*}{\makecell{M3500 [S] \\ (3500, 5453)}}   
        & $F(x)$  & 1123 & 923 & 548 & 407 & \textbf{319} & 1134 & 932 & 553 & 428 & \textbf{336} & 1118 & 917 & 547 & 455 & \textbf{364} \\ 
        & Time    & 575.64 & 480.25 & 185.37 & \textbf{28.64} & 42.93 & 402.87 & 336.09 & 129.82 & \textbf{20.04} & 30.05 & 201.53 & 168.12 & 64.91 & \textbf{10.03} & 15.06 \\ \hline


    \multirow{2}{*}{\makecell{M3500b [S] \\ (3500, 5453)}}  
        & $F(x)$  & $4.66\times10^{4}$ & $3.01\times10^{4}$ & $2.01\times10^{4}$ & $1.43\times10^{4}$ & $\mathbf{1.22\times10^{4}}$ & $4.67\times10^{4}$ & $3.02\times10^{4}$ & $2.02\times10^{4}$ & $1.52\times10^{4}$ & \textbf{$\mathbf{1.31\times10^{4}}$} & $4.66\times10^{4}$ & $3.00\times10^{4}$ & $2.01\times10^{4}$ & $1.63\times10^{4}$ & \textbf{$\mathbf{1.41\times10^{4}}$} \\ 
        & Time    & 578.72 & 482.03 & 188.94 & \textbf{28.97} & 43.52 & 405.11 & 337.44 & 132.19 & \textbf{20.29} & 30.42 & 202.47 & 168.70 & 66.07 & \textbf{10.11} & 15.24 \\ \hline

    \multirow{2}{*}{\makecell{M3500c [S] \\ (3500, 5453)}}  
        & $F(x)$  & $5.64\times10^{4}$ & $4.01\times10^{4}$ & $2.31\times10^{4}$ & $1.62\times10^{4}$ & \textbf{$\mathbf{1.31\times10^{4}}$} & $5.65\times10^{4}$ & $4.02\times10^{4}$ & $2.32\times10^{4}$ & $1.73\times10^{4}$ & \textbf{$\mathbf{1.42\times10^{4}}$} & $5.64\times10^{4}$ & $4.00\times10^{4}$ & $2.30\times10^{4}$ & $1.82\times10^{4}$ & \textbf{$\mathbf{1.51\times10^{4}}$} \\ 
        & Time    & 577.83 & 481.48 & 186.52 & \textbf{28.83} & 42.78 & 404.52 & 337.02 & 130.49 & \textbf{20.23} & 29.96 & 202.17 & 168.51 & 65.29 & \textbf{10.10} & 14.97 \\ \hline

    \multirow{2}{*}{\makecell{City10K [S] \\ (10000, 20687)}} 
        & $F(x)$  & $4.52\times10^{4}$ & $3.51\times10^{4}$ & 5223 & 5073 & \textbf{4012} & $4.53\times10^{4}$ & $3.52\times10^{4}$ & 5235 & 5564 & \textbf{4527} & $4.51\times10^{4}$ & $3.50\times10^{4}$ & 5219 & 6044 & \textbf{5023} \\ 
        & Time    & 950.57 & 810.27 & 250.01 & \textbf{44.97} & 61.23 & 665.35 & 567.19 & 175.04 & \textbf{31.47} & 42.76 & 332.69 & 283.63 & 87.48 & \textbf{15.73} & 21.39 \\ \hline
    \multirow{2}{*}{\makecell{Grid1000 [S] \\ (1000, 1250)}}   
        & $F(x)$  & 3050 & 1800 & 1100 & 950 & \textbf{820}%
                 & 3065 & 1815 & 1120 & 1005 & \textbf{880}%
                 & 3040 & 1790 & 1090 & 1035 & \textbf{910} \\ 
        & Time    & 225.50 & 181.00 & 73.00 & \textbf{13.50} & 19.30%
                 & 157.00 & 128.00 & 51.20 & \textbf{9.80} & 14.40%
                 & 78.20  & 63.00  & 26.00 & \textbf{5.00}  & 7.10 \\ \hline

    \multirow{2}{*}{\makecell{Intel [R] \\ (1228, 1483)}}   
        & $F(x)$  & $9.96\times10^{5}$ & $6.01\times10^{4}$ & 1023 & 517 & \textbf{412} & $9.97\times10^{5}$ & $6.02\times10^{4}$ & 1041 & 531 & \textbf{421} & $9.96\times10^{5}$ & $6.00\times10^{4}$ & 1019 & 549 & \textbf{446} \\ 
        & Time    & 220.40 & 182.26 & 72.53 & \textbf{12.51} & 18.79 & 154.27 & 127.58 & 50.76 & \textbf{8.81} & 13.17 & 77.12 & 63.75 & 25.36 & \textbf{4.43} & 6.55 \\ \hline

    \multirow{2}{*}{\makecell{MIT [R] \\ (808, 827)}}       
        & $F(x)$  & 2048 & 1249 & 1021 & 953 & \textbf{809} & 2049 & 1252 & 1024 & 1003 & \textbf{853} & 2031 & 1241 & 1018 & 1047 & \textbf{892} \\ 
        & Time    & 231.37 & 185.21 & 74.96 & \textbf{14.46} & 20.57 & 161.95 & 129.62 & 52.47 & \textbf{10.05} & 14.38 & 80.98 & 64.81 & 26.19 & \textbf{5.08} & 7.21 \\ \hline
    \multirow{2}{*}{\makecell{CSAIL [R] \\ (1045, 1172)}}   
        & $F(x)$  & 3375 & 2074 & 1138 & 945 & \textbf{802}%
                 & 3387 & 2085 & 1151 & 1005 & \textbf{856}%
                 & 3362 & 2063 & 1139 & 1036 & \textbf{894} \\ 
        & Time    & 235.47 & 187.66 & 75.92 & \textbf{14.78} & 21.07%
                 & 164.24 & 131.43 & 52.93 & \textbf{10.34} & 14.68%
                 & 81.56  & 65.21  & 26.13 & \textbf{5.17}  & 7.27 \\ \hline

  \end{tabular}}
\begin{flushleft}
\footnotesize Note: \textbf{bold} represents best performance. We scale to larger teams (\(n=35\) by replicating actors trained on \(n=7\). [S] and [R] denote \emph{synthetic} and \emph{real-world} datasets, respectively.
Although larger teams might suggest higher inference time, partitioning the same dataset across more robots reduces per-robot workload.
\end{flushleft}
    \vspace{-7mm}
\end{table*}

Under the CTDE regime, GA--SAC maximizes the standard SAC return \cite{iqbal2019actor} as shown:
\vspace{-1mm}
\begin{equation}
    J(\Psi)=
    \sum_{i=1}^{n}\,
    \mathbb{E}_{\tau\sim\Pi}\!
    \Bigl[
        \sum_{t=0}^{T-1}
            \gamma^{t}\bigl(
                R_t^{\,r_i}
                +\alpha\,
                \mathcal{H}\!\bigl[\pi_{\psi^{r_i}}(\cdot\mid s_t^{\,r_i})\bigr]
            \bigr)
    \Bigr],
    \label{eq:sac}
\end{equation}
where $\Psi=\{\psi^{r_1},\dots,\psi^{r_n}\}$ is actors' parameter set,  
$\gamma\!\in\![0,1)$ is the discount factor,  
$\alpha>0$ the temperature, $\mathbb{E}$ is expectation operator, and
$\mathcal{H}$ the Shannon entropy. The policy evaluation, improvement, target networks, and temperature tuning
are unchanged \cite{iqbal2019actor}. Because gradients originate from the central critic but are applied
locally, each robot’s on-board computation scales as
$\mathcal{O}(|\mathcal{E}^{r_i}_t|)$.  
After $T$ steps each robot holds a refined sub-graph
$\mathcal{G}_T^{\,r_i}$. As we preserve inter-robot graph dependencies in graph partitioning, the duplicated separator vertices (e.g, nodes 3 and 4 of $r_1$ in Fig. \ref{fig:intro_1}) are reconciled via the information-weighted ADMM of~\cite{fung2019uncertainty}, exchanging only
separator poses until convergence, after which all sub-graphs merge into a
globally consistent estimate.



\begin{figure}[t]
    \centering
    \includegraphics[width = \linewidth]{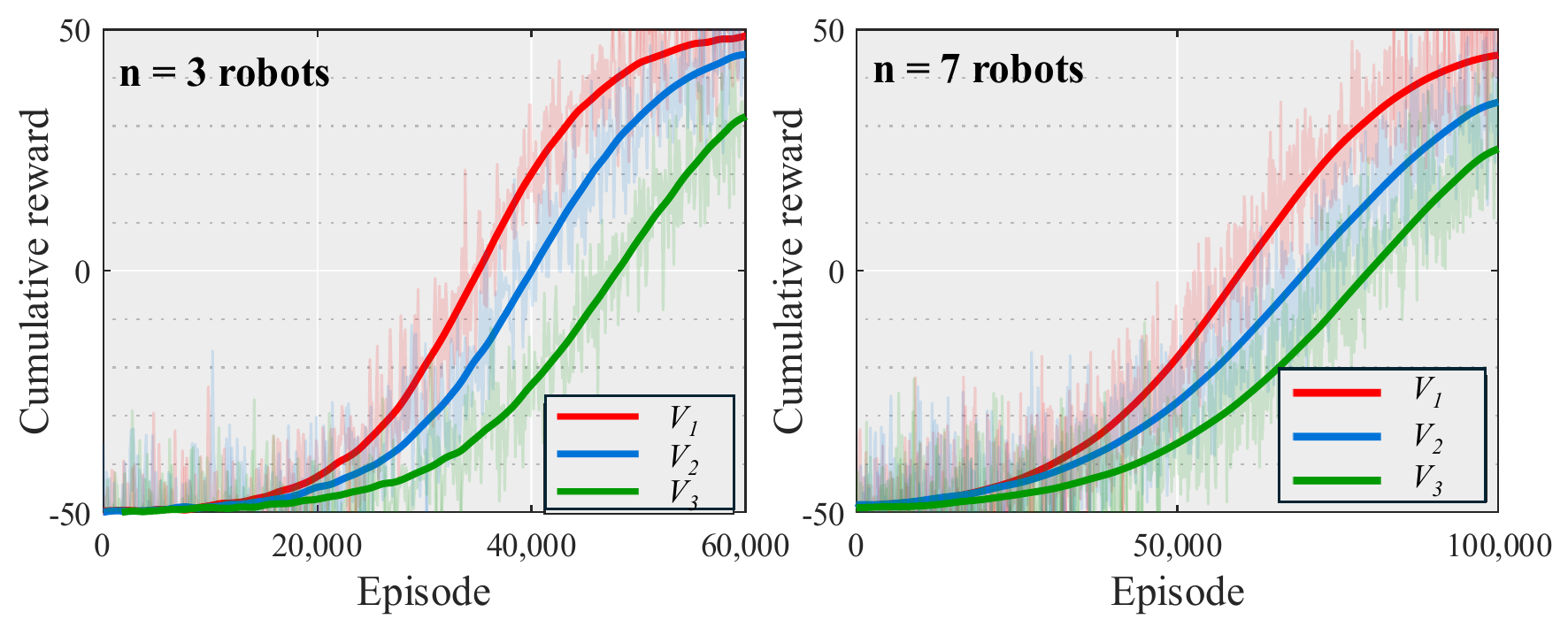}
    \vspace{-5mm}
    \caption{Cumulative  Learning Efficiency of the Proposed Solution Across 3 Training Environments for n=3 (\textbf{Left}) and n=7 (\textbf{Right}) Actors.}
    \label{fig:learningcurves}
    \vspace{-4mm}
\end{figure} 

\begin{figure}[b]
    \vspace{-5mm}
    \centering
    \includegraphics[width = \linewidth]{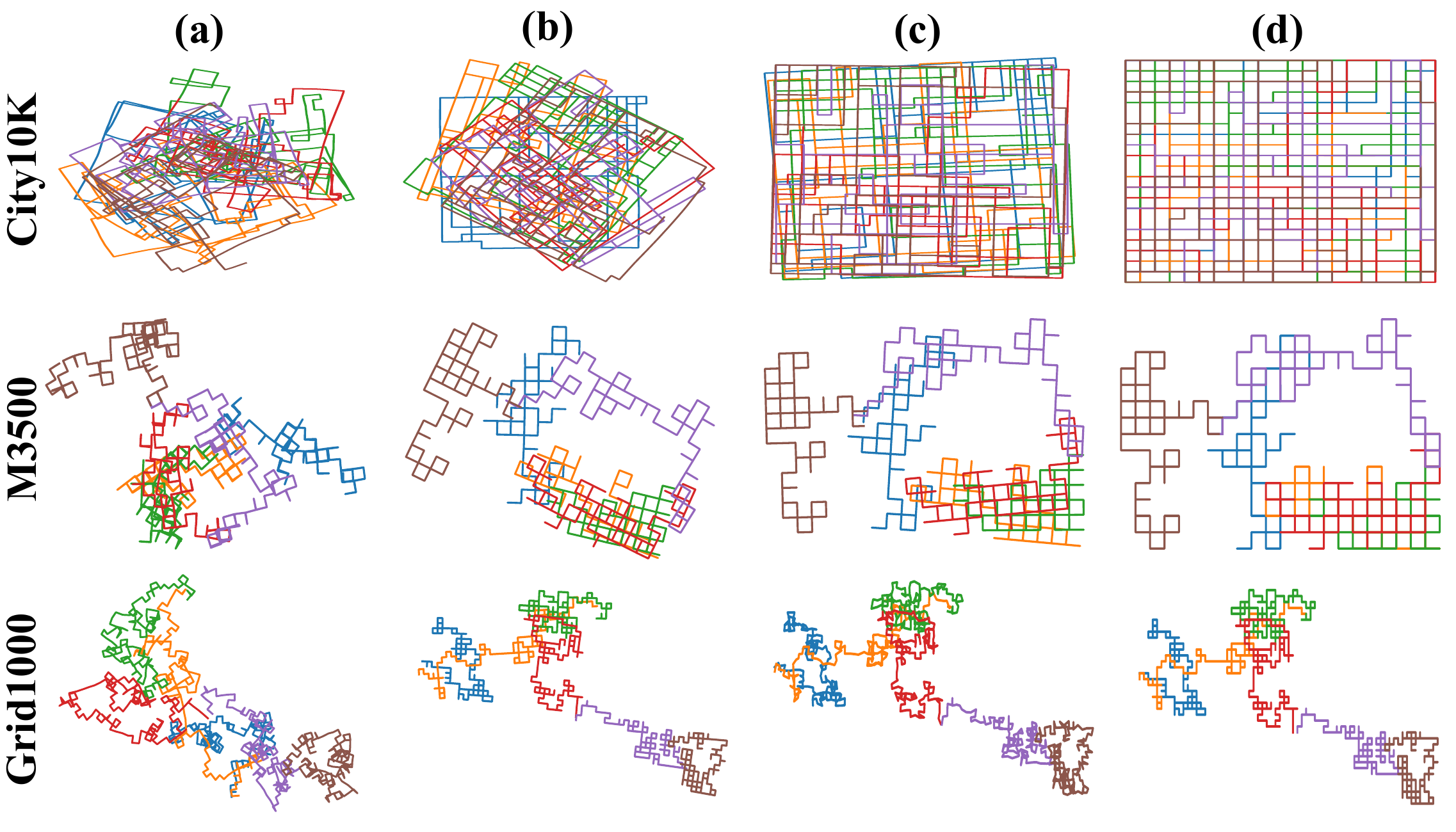}
    \vspace{-8mm}
    \caption{(a) Initial Noisy Estimate (b) MM-PGO (c) Prop-V1 (d) Prop-V2 results on datasets, where each color represents the trajectory of each robot.}
    \label{fig:result_pg}
    \end{figure} 

\section{Experimental Results \& Analysis}
Due to its inherently distributed architecture, our framework tackles a fundamentally different problem from single-agent RL-PGO~\cite{kourtzanidis2023rl} and is therefore not directly comparable. Instead, we compare our method against the state-of-the-art distributed frameworks: CC-PGO \cite{tian2021distributed}, ML-PGO \cite{li2024distributed}, and MM-PGO \cite{fan2023majorization}. 
We adopt the canonical benchmark suite \cite{lucacarlone-datasets} (introduced in MM-PGO \cite{fan2023majorization}) and two additional datasets (Grid1000 and City10K) from \cite{planarpgo-datasets}, providing a diverse list of synthetic and real-world dataset variants.
For fairness, all baseline frameworks are executed for 1000 iterations. Throughout the paper, we refer to the pure learned solver as Prop-V1, and to its bootstrapped counterpart—where the learned correction of the local pose graph is followed by 75 iterations of a classical Levenberg–Marquardt (LM) refinement—as Prop-V2. While Prop-V1 already delivers near-optimal pose estimates, integrating 75 Levenberg–Marquardt iterations in Prop-V2 secures provable convergence, adding certifiability, numerical stability, and seamless drop-in compatibility with SLAM stacks. We evaluate performance using the global objective value \(F(x)\) defined in \eqref{eq:local_pgo} and the inference time as the metrics for comparison.

\subsection{Training Environments \& Learning}
For each training episode, we synthesise noise–free ground-truth trajectories and drifted initial estimates for teams of $n=3-7$ robots, with each robot contributing $60$ poses, with a loop-closure sampling ratio of 0.15. The measurement uncertainty is modeled as isotropic $\mathbb{SE}(2)$ noise, with distinct standard deviations applied respectively to odometry, intra-robot loop closures, inter-robot relative estimates and inter-robot loop closures across three training environments: $V_1$=$\{0.06, 0.10, 0.14\}$, $V_2$=$\{0.10, 0.14, 0.18\}$, and $V_3$=$\{0.14, 0.18, 0.22\}$. Each episode lasts for a number of steps equal to the poses in the local graph to ensure that all poses are iteratively corrected. Transitions are discounted with $\gamma\!=\!0.99$ and stored in a $30{,}000$-entry replay buffer; mini-batches of $64$ are sampled after a $10{,}000$-step warm-up, and soft target updates ($10^{-3}$) are applied at every step. Actors and the central critic are trained with Adam (learning rates $3\times10^{-4}$ and $3\times10^{-5}$) respectively. Each actor comprises a 5-layer edge-conditioned GNN (hidden size $128$), a Gumbel-softmax edge selector (that expands $128$ feature vector to $200$ dimensions), and a 2-layer MLP (hidden size $512$) predicting pose corrections.  The critic concatenates edge indices and pose corrections and feeds the resulting 512-D vector into a 2-layer MLP (hidden size: 512 and 256) that returns a scalar value. Training was carried out on a CUDA-enabled Intel i9 workstation (32GB RAM) equipped with an NVIDIA RTX 4080 GPU (16GB RAM), whereas all inference experiments were executed purely on the CPU.

The cumulative learning curves in Fig. \ref{fig:learningcurves} show that lower measurement noise ($V_1$) leads to faster convergence and higher cumulative rewards for both team sizes, while increasing the number of robots requires more training episodes to achieve similar performance. Additionally, the comparative performance of the proposed solutions (trained in $V_2$ environment) against existing state-of-the-art benchmarks is detailed in Table~\ref{tab:ale_time_benchmarks}, with sample pose graph optimization results for the proposed methods shown in Fig. \ref{fig:result_pg}.  Overall, Prop-V1 reduces the residual global objective \(F(x)\) by 25\%–50\% relative to existing baselines, and achieves inference-time speed-ups of 6.5× over MM-PGO and 20.1× over CC-PGO. Prop-V2 further decreases \(F(x)\) by an additional 15\%–25\%, with inference times increasing by approximately 50\% with respect to Prop-V1 due to the inclusion of the LM refinement. These results unequivocally demonstrate that our approach outperforms state-of-the-art baselines in both efficiency and final accuracy. Moreover, the Prop-V2 outcomes confirm that Prop-V1 provides a powerful initialization for classical solvers, effectively bootstrapping the subsequent refinement process.


\subsection{Denoising Robustness on Real-World Benchmarks}
\label{ssec:aeg_realworld}
We evaluate the AEG denoising module under realistic conditions by injecting controlled outliers directly into standard real-world pose-graph datasets. For each trial, we randomly corrupt a fraction of all loop-closure and inter-robot estimate constraints,
replacing the true relative transform with a flipped rotation $\theta\!\sim\!\mathcal U(-\pi,\pi)$
and a perturbed translation $\mathbf t\!\sim\!\mathcal N(\mathbf 0,0.5^2\mathbf L_{avg})\,\text{m}$, where $L_{avg}$ denotes the average translation magnitude in the dataset. We subsequently optimize each corrupted graph using the proposed approach and MM-PGO augmented with its kernel-based denoising scheme, as shown in Table \ref{tab:real_outlier} with some representative datasets. The evaluation metrics analyzed include the outlier precision and recall. 

\begin{table}[b]
\centering
\vspace{-2mm}
\caption{Performance comparison of denoising modules}
\vspace{-1mm}
\label{tab:real_outlier}
\scriptsize
\setlength{\tabcolsep}{2pt}
\renewcommand{\arraystretch}{1.05}
\resizebox{\columnwidth}{!}{%
\begin{tabular}{|l|c|cc|cc|cc|}
\hline
\multirow{2}{*}{Dataset} & \multirow{2}{*}{\%} &
\multicolumn{2}{c|}{\(F(x)\)} &
\multicolumn{2}{c|}{Precision} &
\multicolumn{2}{c|}{Recall} \\ \cline{3-8}
& & Prop-V1 & MM-PGO & Prop-V1 & MM-PGO & Prop-V1 & MM-PGO \\ \hline
\multirow{3}{*}{Intel}
  & 2.5\% & \(\mathbf{5.6\times10^{2}}\) & \(1.1\times10^{3}\) & \(\mathbf{0.93}\) & 0.81 & \(\mathbf{0.85}\) & 0.66 \\
  & 5\%    & \(\mathbf{8.4\times10^{2}}\) & \(1.9\times10^{3}\) & \(\mathbf{0.91}\) & 0.77 & \(\mathbf{0.82}\) & 0.60 \\
  & 10\%   & \(\mathbf{1.6\times10^{3}}\) & \(3.4\times10^{3}\) & \(\mathbf{0.88}\) & 0.69 & \(\mathbf{0.79}\) & 0.55 \\ \hline
\multirow{3}{*}{MIT}
  & 2.5\%  & \(8.1\times10^{2}\) & \(\mathbf{7.9\times10^{2}}\) & \(\mathbf{0.92}\) & 0.80 & \(\mathbf{0.84}\) & 0.64 \\
  & 5\%    & \(\mathbf{1.1\times10^{3}}\) & \(2.3\times10^{3}\) & 0.90 & \(\mathbf{0.91}\) & \(\mathbf{0.82}\) & 0.58 \\
  & 10\%   & \(\mathbf{2.0\times10^{3}}\) & \(4.1\times10^{3}\) & \(\mathbf{0.87}\) & 0.67 & \(\mathbf{0.78}\) & 0.53 \\ \hline
\multirow{3}{*}{CSAIL}
  & 2.5\%  & \(\mathbf{0.6\times10^{3}}\) & \(1.1\times10^{3}\) & \(\mathbf{0.94}\) & 0.82 & 0.86 & \(\mathbf{0.87}\) \\
  & 5\%    & \(\mathbf{1.2\times10^{3}}\) & \(2.1\times10^{3}\) & \(\mathbf{0.92}\) & 0.79 & \(\mathbf{0.83}\) & 0.63 \\
  & 10\%   & \(\mathbf{3.6\times10^{3}}\) & \(6.4\times10^{3}\) & \(\mathbf{0.89}\) & 0.72 & \(\mathbf{0.80}\) & 0.57 \\ \hline
\end{tabular}}
\begin{flushleft}
\footnotesize Note: “\%” denotes the percentage of edges that are intentionally corrupted and \textbf{bold} represents the method that outperformed. CC-PGO and ML-PGO are omitted, as they lack explicit denoising mechanisms.
\end{flushleft}
\end{table}

Across the three datasets, Prop-V1 consistently drives down the global objective \(F(x)\), achieving a median 38\,\% reduction relative to MM-PGO and outperforming the baseline in eight of the nine corruption scenarios, with the most pronounced gain—a 47\,\% drop—registered on CSAIL at 10\,\% outlier level.  In terms of robustness, Prop-V1 maintains high precision in the 0.88–0.94 range, resulting in an average absolute improvement of 0.13 over MM-PGO and indicating a more selective pruning of erroneous constraints.  Moreover, Prop-V1 retains a larger share of true inliers, delivering a mean recall advantage of 0.19 across all settings and reaching 0.85 on the Intel dataset. These improvements in precision and recall arise without sacrificing either metric, underscoring a superior precision–recall trade-off that persists as noise intensifies.  Collectively, the results confirm that the AEG denoiser not only tightens residuals but also sustains reliable data association, positioning Prop-V1 ahead of the MM-PGO.

\begin{figure}[t]
    \centering
    \includegraphics[width = \linewidth]{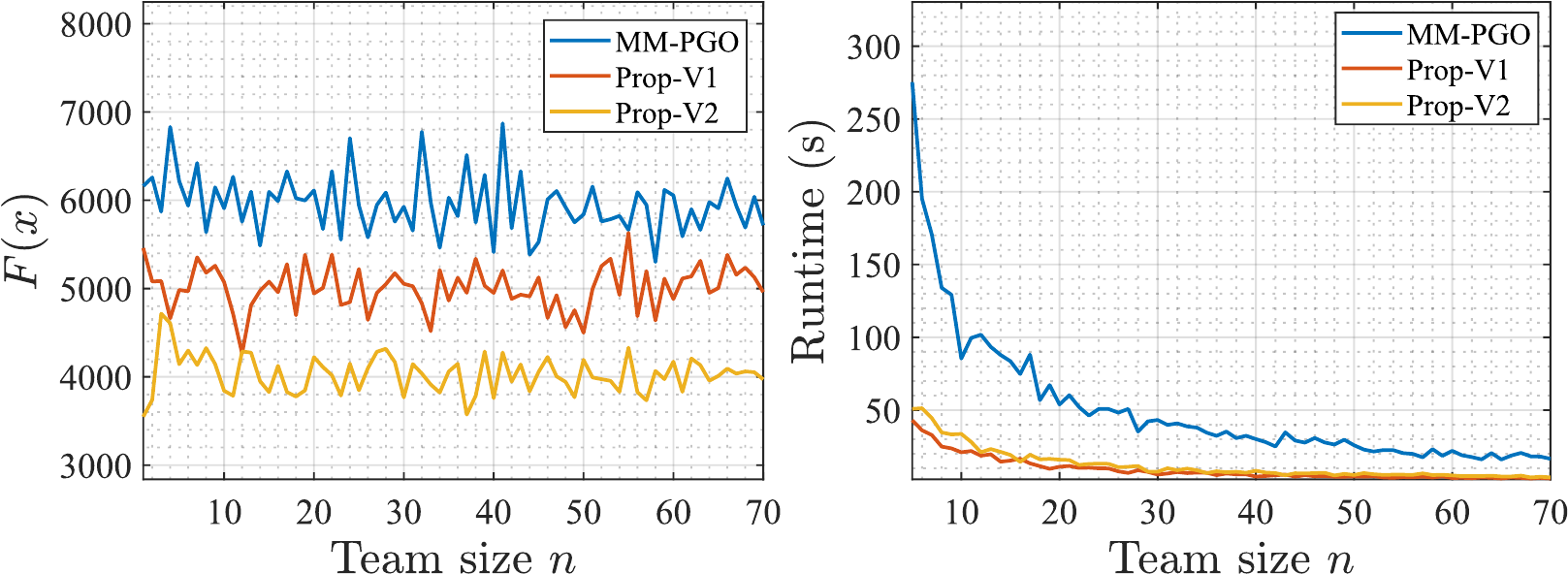}
    \vspace{-7mm}
    \caption{\textbf{(Left)} Global objective $F(x)$ \textbf{(Right)} Runtime versus team size \(n\).}
    \label{fig:scalability}
    \vspace{-5mm}
\end{figure} 

\vspace{-1mm}
\subsection{Scalability: Actor replication}
To evaluate scalability, we reuse the actors trained for the \(n=7\) configuration and replicate them—without any retraining—to assemble larger teams, generating the performance profile for \(n=1\) to \(70\) robots shown in Fig.~\ref{fig:scalability}. This replication is feasible because our distributed policies are explicitly designed to be agent-agnostic: each policy operates solely on local observations and the structure of its immediate subgraph, without relying on global system variables such as the total number of robots, team composition, or graph size. Across the entire range of team sizes, both learned methods deliver
substantially lower objective values than the classical
MM-PGO solver.  Averaged over all \(n\), Prop-V1
reduces \(F(x)\) by roughly \(15\,\%\) relative to MM-PGO,
while Prop-V2 yields a further \(15–20\,\%\) drop,
culminating in a \(32.5\,\%\) improvement over the baseline. Although \textsc{MM-PGO} displays the expected
\(O(\tfrac{1}{n})\) runtime trend, it remains one to two orders of
magnitude slower than our methods.  For small teams
(\(n\le 7\)), MM-PGO exceeds \(250\,\mathrm{s}\), whereas Prop-V2 solves in \(<\!40\,\mathrm{s}\).  Beyond
\(n\ge 20\), Prop-V2 stabilises below \(10\,\mathrm{s}\), providing a consistent \(10–15\times\) speed-up over MM-PGO and about \(2\times\) over Prop-V1. These results empirically validate that actor replication enables seamless scalability, demonstrating that a single set of learned, agent-agnostic policies can generalize across team sizes while consistently outperforming classical solvers.

\vspace{-1mm}
\section{Conclusion}
We introduced a MARL-based distributed 2D pose-graph optimizer that fuses edge-conditioned GNNs with adaptive gating to simultaneously denoise local sub-graphs and produce consistent pose estimates via edge-wise pose corrections. Comprehensive evaluations on diverse synthetic and real-world benchmarks show that our approach surpasses existing baselines in accuracy, outlier robustness, and scalability, operating both as an efficient stand-alone solver and as an initializer that accelerates downstream optimization. Moreover, we can also integrate the proposed solution with camera-pose estimation within structure-from-motion pipelines, aiming to produce high-quality pose estimates that further accelerate bundle adjustment while preserving reconstruction accuracy.

\bibliographystyle{IEEEtran}
\bibliography{Main}

\end{document}